\documentclass[runningheads]{llncs}

\usepackage[mobile]{eccv}

\usepackage{eccvabbrv}

\usepackage{graphicx}
\usepackage{booktabs}
\usepackage{marvosym}

\usepackage[accsupp]{axessibility}  

\usepackage{hyperref}
\hypersetup{pagebackref,breaklinks,colorlinks,citecolor=eccvblue}

\usepackage{orcidlink}

\usepackage[font=small,labelfont=bf,tableposition=top]{caption}
\DeclareCaptionLabelFormat{andtable}{#1~#2  \&  \tablename~\thetable}
\usepackage{wrapfig}

\begin{document}

\title{Motion-Oriented Compositional Neural Radiance Fields for Monocular Dynamic Human Modeling} 

\titlerunning{MoCo-NeRF}

\author{Jaehyeok Kim\inst{1} \and
Dongyoon Wee\inst{2} \and
Dan Xu\inst{1}$^{\href{danxu@cse.ust.hk}{\textcolor{black}{\textrm{\Letter}}}}$}

\renewcommand{\thefootnote}{}
\footnotetext[1]{$^{\href{danxu@cse.ust.hk}{\textcolor{black}{\textrm{\Letter}}}}$ Corresponding author.}

\authorrunning{J.~Kim et al.}

\institute{\textsuperscript{1} HKUST, Hong Kong \qquad
\textsuperscript{2} NAVER Cloud Corp., South Korea\\
\email{jkimbf@cse.ust.hk}, \email{dongyoon.wee@navercorp.com}, \email{danxu@cse.ust.hk}}

\maketitle

\begin{abstract}
    This paper introduces \textbf{Mo}tion-oriented \textbf{Co}mpositional \textbf{Ne}u-ral \textbf{R}adiance \textbf{F}ields (MoCo-NeRF), a framework designed to perform free-viewpoint rendering of monocular human videos via novel non-rigid motion modeling approach. In the context of dynamic clothed humans, complex cloth dynamics generate non-rigid motions that are intrinsically distinct from skeletal articulations and critically important for the rendering quality. The conventional approach models non-rigid motions as spatial (3D) deviations in addition to skeletal transformations. However, it is either time-consuming or challenging to achieve optimal quality due to its high learning complexity without a direct supervision. To target this problem, we propose a novel approach of modeling non-rigid motions as radiance residual fields to benefit from more direct color supervision in the rendering and utilize the rigid radiance fields as a prior to reduce the complexity of the learning process. Our approach utilizes a single multiresolution hash encoding (MHE) to concurrently learn the canonical T-pose representation from rigid skeletal motions and the radiance residual field for non-rigid motions. Additionally, to further improve both training efficiency and usability, we extend MoCo-NeRF to support simultaneous training of multiple subjects within a single framework, thanks to our effective design for modeling non-rigid motions. This scalability is achieved through the integration of a global MHE and learnable identity codes in addition to multiple local MHEs. We present extensive results on ZJU-MoCap and MonoCap, clearly demonstrating state-of-the-art performance in both single- and multi-subject settings.~The code and model will be made publicly available at the project page: \href{https://stevejaehyeok.github.io/publications/moco-nerf}{https://stevejaehyeok.github.io/publications/moco-nerf}.
\end{abstract}

\begin{figure}[!t]
    \centering
    \includegraphics[width=.99\textwidth]{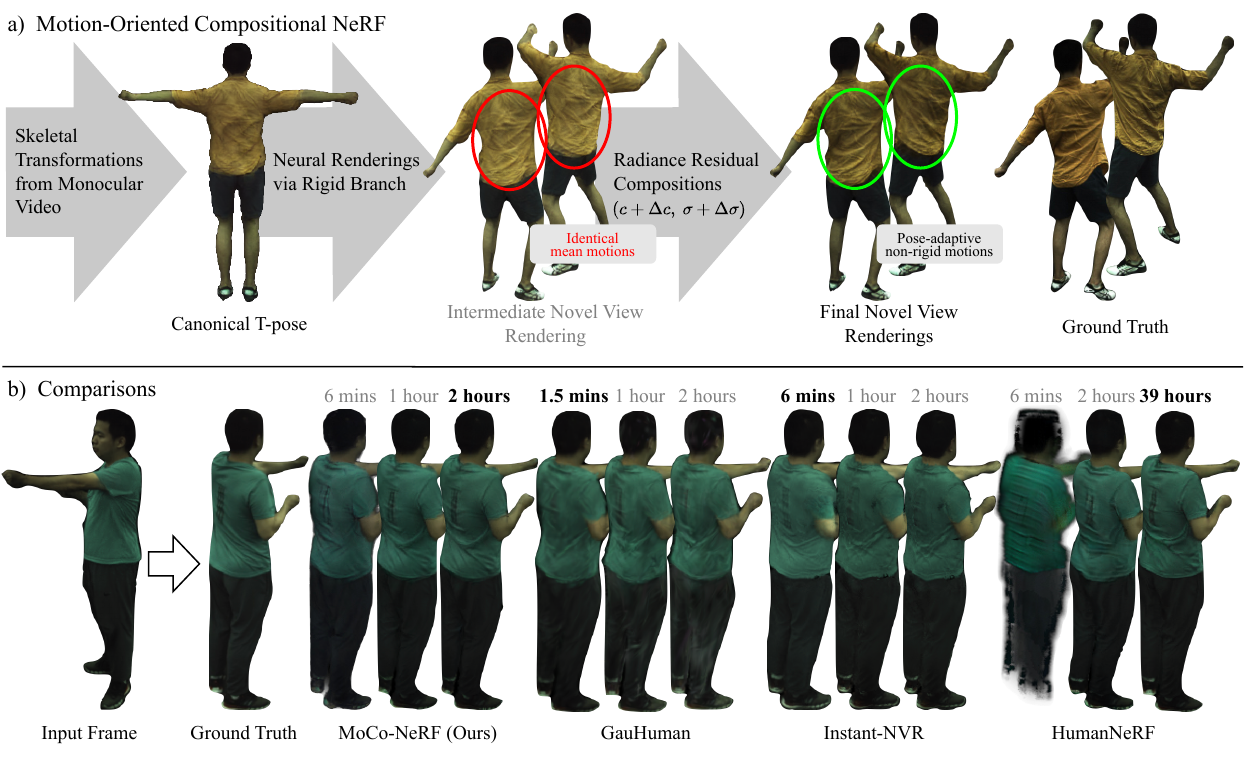} 
    \caption{
    a) We introduce a motion-oriented compositional NeRF for photo-realistic modeling of dynamic humans from monocular videos. Our approach innovatively employs radiance compositions to capture pose-adaptive non-rigid motions, overcoming the limitations of skeletal transformations that typically yield an average of observed deformations (mean motions).
    b) The proposed MoCo-NeRF achieves state-of-the-art rendering quality and noteworthy efficiency in novel view synthesis, compared to leading competitors~\cite{hu2024gauhuman, geng2023instantnvr, weng2022humannerf}. The \textbf{bold} denotes the proposed training duration of each comparison model. 
    MoCo-NeRF uniquely captures coherent non-rigid motions, like T-shirt wrinkles relative to body pose, from entirely new viewpoints. Moreover, MoCo-NeRF significantly surpasses the efficiency of another photo-realistic method, HumanNeRF~\cite{weng2022humannerf}.}
    \label{fig:teaser}
\end{figure}

\section{Introduction}
\label{sec:intro}

Novel view synthesis of monocular human video with photo-realistic quality is a challenging problem, as it involves complex articulated body motions and necessitates sophisticated modeling of non-skeletal or non-rigid motions (\eg, wrinkles of clothes) from unseen viewpoints. In particular, non-rigid motion modeling is essential for achieving photo-realism and coherence in free-viewpoint renderings. A prevalent solution adopted by recent methods~\cite{weng2022humannerf, geng2023instantnvr, hu2024gauhuman} models the non-rigid motions as pose-dependent 3D spatial offsets of canonical point samples. However, the offset-based strategies lead to either extensive training time or irrelevant modeling of non-rigid motions due to its high learning complexity, incurred by the following reasons. \emph{First}, a single spatial offset spans an unbounded range and is unique for every 3D point across all input frames and body poses, thereby escalating computational complexity. \emph{Second}, the learning objective of spatial offsets is not directly aligned with the conventional RGB-based supervision, further complicating the modeling process.
\par To target the issues above-discussed, in this paper, we propose Motion-oriented Compositional Neural Radiance Fields (MoCo-NeRF), a new framework that effectively \emph{models non-rigid motions at a radiance-level through a unique design of residual radiance field learning} for free-viewpoint renderings of monocular human videos.~Specifically, our proposed framework decomposes a single MHE~\cite{mueller2022instant} into two branches: a rigid branch that learns a canonical T-pose radiance field solely via rigid skeletal transformations, and a non-rigid branch that optimizes pose-driven \emph{radiance residuals} for capturing appearance discrepancies between renderings of the rigid branch and the ground-truth images, which are typically caused by non-rigid motions. 
The proposed approach circumvents the complexities associated with learning unbounded spatial offsets by reducing the prediction range to the radiance space with the rigid radiance field as a prior. It further establishes a more direct supervision from the ground-truth pixel colors for optimizing the rigid and non-rigid radiance fields by embedding them together in the rendering process, which can further enhance the learning efficiency.   
Besides, we present a cross-attention module to query a pose conditioned implicit feature from a learnable base code. This process allows our framework to extract more discriminative implicit features tailored to each particular pose, helping our non-rigid branch to learn coherent pose-driven radiance residuals.

In addition, thanks to our effective design for modeling non-rigid motions, the proposed MoCo-NeRF architecture can be flexibly extended to support unified training across multiple subjects while maintaining photo-realistic quality, without a significant increase in training time. For this architectural extension, we introduce multiple local MHEs and a global MHE, and also extend the aforementioned base code to a set of identity (ID) codes as a learnable codebook.
The multiple local MHEs are designed for modeling multiple subjects and the global MHE targets the learning of generic and universal body representations.
Furthermore, subject-specific learnable ID codes that are pivotal for generating pose-embedded implicit features enable effective sharing of the MLP decoder in the rigid branch by all subjects, enhancing the model's efficiency and scalability.

\par MoCo-NeRF achieves state-of-the-art performance on free-viewpoint rendering of \emph{monocular} human videos in both single- and multi-subject settings, showing clear improvements compared to the prior works. We present our extensive results on ZJU-MoCap~\cite{peng2021neuralbody} and MonoCap~\cite{peng2021animatable, habermann2021, habermann2020deepcap} to demonstrate the effectiveness of the proposed MoCo-NeRF in learning photo-realistic representation and modeling complex non-rigid motions from monocular videos.
\par In summary, the contribution of our work is threefold:
\begin{itemize}
    \item 
    We propose a novel motion-oriented compositional NeRF approach, MoCo-NeRF, which concurrently learns canonical rigid radiance fields and pose-dependent radiance residual fields, effectively handling both rigid and non-rigid motions.
    We also introduce a pose-embedded implicit feature, computed by cross-attention between a body pose and a designed learnable base code, to implicitly enhance pose-dependent radiance residual learning.
    \item 
    The proposed MoCo-NeRF is the first work that can be readily extended to enable unified training of multiple subjects without significantly increasing the GPU memory demands.
    We achieve this through our local MHEs, a global MHE, and the extended identity codebook, which enable subject-discriminative representation learning and effective module sharing.
    \item 
    The proposed MoCo-NeRF achieves state-of-the-art performance under both single- and multi-subject training settings with a significant efficiency improvement compared to the best-performing methods. MoCo-NeRF jointly learns all 6 subjects of ZJU-MoCap on one RTX 3090 GPU within 2 hours. 
\end{itemize}

\section{Related Work}
We review closely related work from three perspectives: deformable neural rendering, neural human rendering, and monocular photo-realistic human rendering.
\par\noindent\textbf{Deformable neural rendering.} NeRF~\cite{mildenhall2020nerf} leverages an MLP to learn a static 3D representation of a scene from a dense set of images from diverse viewpoints. 
For its application in real-life scenarios, prior works have enhanced NeRF in various directions such as improving efficiencies~\cite{garbin2021fastnerf, reiser2021kilonerf, yu2021plenoctrees, hedman2021snerg, yu_and_fridovichkeil2021plenoxels, mueller2022instant, chen2022tensorf, liu2020neuralsparse} or relieving strong reliance on impractical settings like dense input views~\cite{yu2020pixelnerf, jain2021dietnerf, chen2021mvsnerf, wang2021ibrnet, liu2022neuray}. 
In addition, there are more branches of work such as large-scale modeling~\cite{xiangli2022bungeenerf, turki2022meganerf, tancik2022blocknerf, xu2023gridguided, rematas2022urf}, improving rendering quality~\cite{barron2022mipnerf360, barron2021mipnerf}, reconstruction quality~\cite{oechsle2021unisurf, wang2021neus, yariv2021volume} and more.
However, they are restricted to static scenes while the majority of the real-world objects are dynamic.

Recent works including~\cite{gao2021dynerf, li2020neuralflow, pumarola2020dnerf, park2021nerfies, tretschk2021nonrigid, park2021hypernerf, xian2021spacetime, du2021nerflow} have broadened the modeling capabilities of NeRF to dynamic scenes with small and simple movements/deformations. 
However, they often fail in handling more dynamic or complex motions like human body articulations.
In contrast, our approach enables learning of human-articulated motions from monocular videos, and performing free-viewpoint rendering of a human performer at any time frame of the videos.

\par\noindent
\textbf{Neural rendering of humans.} 
Many early-stage prior works for neural human renderings utilize traditional techniques like multi-view stereo~\cite{guo2019relightables, schonberger2016sfm, johannes2016mvs, wu2020multi} or to build sophisticated systems for capturing human motions~\cite{collet2015hqstream, dou2016fusion4d, su2020robustfusion, martin-brualla2018lookinggood}. However, constructing a dense set of training views and building those capturing systems are expensive, making them unsuitable for real-life situations. \cite{alldieck2022phorhum, saito2019pifu, saito2020pifuhd, chen2022gdna} address 3D human reconstruction by utilizing ground truth human scans, relieving the aforementioned requirements. \cite{dong2023ag3d, hong2022eva3d, noguchi2022unsupervised} further improves it by enabling human reconstruction from 2D image collections. In recent days, there have been many works learning an implicit human representation for novel pose synthesis~\cite{hu2022hvtr, jiang2022selfrecon, li2022tava, peng2021animatable, remelli2022drivable, wang2022arah, zheng2022structured, zhi2022dualspacenerf, huang2022elicit, jiang2022neuman, su2021anerf, zheng2023avatarrex} or for a novel-view synthesis~\cite{kwon2021neural, liu2021neural, peng2021neuralbody, weng2022humannerf, zhao2022humannerf, xu2021hnerf, zhang2022ndf, su2021anerf, jiang2022neuman, jiang2023instantnvr2, weng2023personnerf, peng2024animatable}. There also have been several generalizable works~\cite{mihajlovic2022keypointnerf, shao2022doublefield, mu2023actorsnerf} for a novel human subject with pre-training on multi-view dataset and a few-shot fine-tuning. \cite{chen2023uv, xu20244k4d} focus on improving rendering efficiency to achieve real-time free-view rendering of dynamic humans. Furthermore, \cite{geng2023instantnvr, jiang2023instantnvr2, jiang2022instantavatar, peng2022selfnerf} adopt explicit hash encoding~\cite{mueller2022instant} in order to boost the training efficiency for human representation modeling. In contrast to our problem setup, most of those works require multi-view videos or do not model the non-rigid motions.

\par\noindent\textbf{Monocular photo-realistic human rendering.} HumanNeRF~\cite{weng2022humannerf} presents a state-of-the-art performance by optimizing motion fields and canonical T-pose representation of a human performer. However, its implicit modeling of pose-dependent non-rigid motions requires tens of hours to converge. Instant-NVR~\cite{geng2023instantnvr} achieves efficient training with part-based representation and 2D motion parameterization for learning non-rigid motions as 3D offsets. Note that the human parametric model (SMPL)~\cite{loper2015smpl}, an extra requirement for training, plays a significant role in their framework. In addition, a concurrent work GauHuman~\cite{hu2024gauhuman}, which also employs spatial offset approach with Gaussian Splatting (GS)~\cite{kerbl3Dgaussians}, further enhances the efficiency and performance upon \cite{geng2023instantnvr} by leveraging SMPL for the density control of GS. Nevertheless, \cite{geng2023instantnvr, hu2024gauhuman} struggle to learn complex non-rigid motions, even with an extended training duration as shown in Fig.~\ref{fig:teaser}. In this paper, instead of directly learning spatial offsets like these existing works, we propose a novel approach that effectively models non-rigid body motions by learning motion-oriented compositional radiance fields conditioned by pose-embedded implicit features. Our approach clearly outperforms the best-performing competitors HumanNeRF~\cite{weng2022humannerf}, Instant-NVR~\cite{geng2023instantnvr}, and GauHuman~\cite{hu2024gauhuman} and can further extend to support joint training of multiple subjects.

\begin{figure*}[!t]
\centering
  \includegraphics[width=1\linewidth]{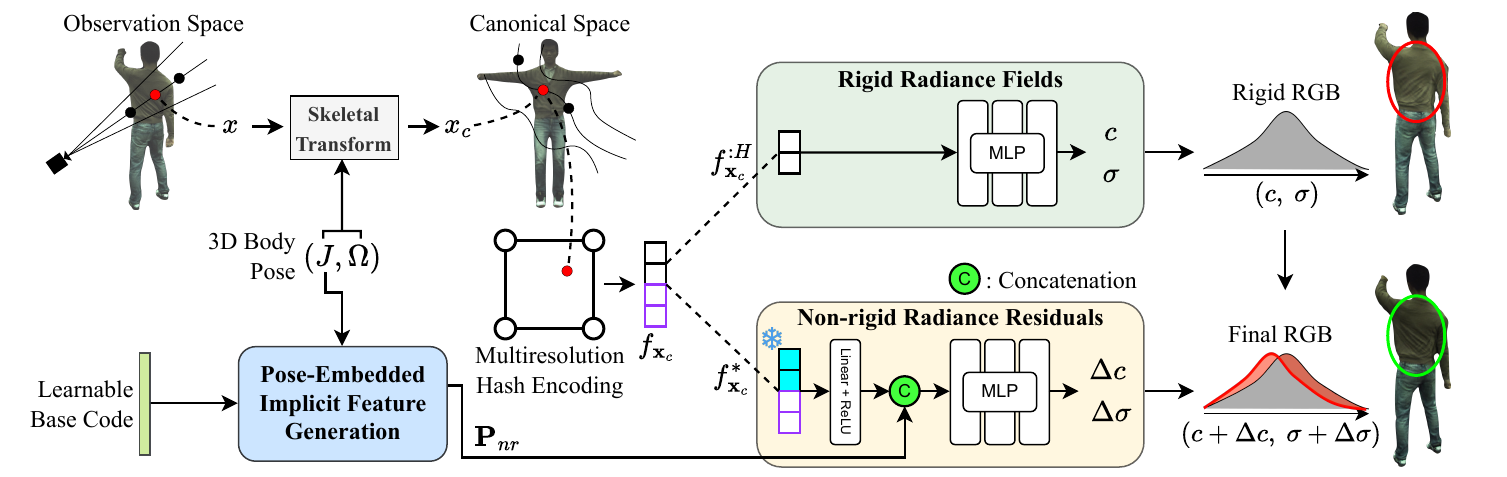}%
  \caption{Overview of the proposed framework MoCo-NeRF for free-view human rendering from a monocular video. Without estimating a geometrical offset of each continuous canonical point for each body pose, our framework is able to handle all deformations via its radiance-compositional approach with a single MHE~\cite{mueller2022instant} and achieve state-of-the-art performance. The pose-embedded implicit feature further enhances the learning of non-rigid radiance residuals by enabling pose-distinctive representation learning.
  }
  \label{fig:overview}
\end{figure*}

\section{The Proposed MoCo-NeRF Method}
The proposed MoCo-NeRF aims to effectively learn high-fidelity representations for free-viewpoint rendering of dynamic human subjects in monocular videos.
We introduce our novel radiance compositional NeRF approach based on different motion types: rigid and non-rigid motions, as illustrated in Fig.~\ref{fig:overview}. The composition of canonical radiance and the per-sample pose-dependent radiance residuals (Sec.~\ref{sec:method}) facilitates photo-realistic quality of rendering with fine-grained modeling of non-rigid motions, rather than leveraging the conventional spatial 3D offset learning approach. 
We further improve the quality of non-rigid radiance residuals by introducing a pose-embedded implicit features~(Sec.~\ref{method:poseCodes}) that is jointly optimized with the non-rigid radiance branch. 
In Sec.~\ref{method:multiSubjects}, thanks to our effective design of non-rigid motion modeling, it enables MoCo-NeRF to be flexibly extended for unified training of multiple subjects with a high efficiency.

\subsection{Preliminary: Skeletal Transformations}\label{sec:skel}
The skeletal motion learns an inverse formulation of the linear blend skinning~\cite{lewis2000inverseLBS}, which transforms canonical points to different body poses by skinning weights.
Weng~\etal~\cite{weng2022humannerf} reformulate the linear blend skinning to transform a point $\mathbf{x}$ in the observation space to a canonical point $\mathbf{x}_c$:
\begin{equation}
    \textbf{x}_c = \displaystyle\sum_{k=1}^K w^o_{k}(R_k\textbf{x} + \textbf{t}_k),
    w^o_{k} = \frac{w^c_k(R_k\textbf{x} + \textbf{t}_k)}{\textstyle\sum_{j=1}^K w^c_{j}(R_j\textbf{x} + \textbf{t}_j)}, \label{eq:1}
\end{equation}
where $w^o_{k}$ and $w^c_{k}$ respectively represent observation and canonical skinning weights of bone $k$, and $R_k$ and $\mathbf{t}_k$ are the rotation and translation of bone $k$, respectively.
Given Eq.~\ref{eq:1}, following~\cite{weng2022humannerf}, we optimize an explicit volume decoder to optimize $\{w^c_k\}^K_{k=1}$ where $K$ is the total number of joints.

\subsection{Motion-Oriented Compositional Neural Radiance Fields}\label{sec:method}
\subsubsection{Rigid neural branch for learning canonical radiance field.} 
The rigid branch of MoCo-NeRF is specifically designed to learn the canonical T-pose representation of the target subject using only rigid body movements.
Our approach is different from but simpler than the prevalent spatial offset learning approach, which incorporates complex modeling of non-rigid motions as $\Delta \mathbf{x}_c$.
Our rigid branch instead directly predicts the color and density of the canonical point $\mathbf{x}_c$ transformed by inverse linear blend skinning (Sec.~\ref{sec:skel}), as depicted in Fig.~\ref{fig:overview}.
First, we query multiresolution volumetric features $f_{\mathbf{x}_c}$ of the canonical point $\mathbf{x}_c$ from the MHE $\psi_{hash}$ as follows: 
\begin{equation}
    f_{\mathbf{x}_c} = \psi_{hash}(\mathbf{x}_c).
\end{equation}
Then, we slice the first half ($H$ dimensions) of $f_{\mathbf{x}_c}$ at each level of resolution to compute $f^{:H}_{\mathbf{x}_c}$ and predicts the radiance $c$ and density $\sigma$ of the rigid canonical point $\mathbf{x}_c$ with an MLP decoder $\mathbf{M}_{rigid}$ as follows:
\begin{equation}
    c,\ \sigma = \mathbf{M}_{rigid}(f^{:H}_{\mathbf{x}_c}).
\end{equation}

\noindent
\textbf{Non-rigid neural branch for learning radiance residual field.}
While the canonical representation does not explicitly model pose-dependent non-rigid motions, it still captures an average of such motions (mean motions) after training. However, the mean motions cannot effectively reflect the complex and realistic non-rigid motions under different body poses. It is only capable of rendering identical deformations across different poses, as can be observed in Fig.~\ref{fig:teaser}.
To address this effectively, the non-rigid branch of MoCo-NeRF is designed to concurrently optimize for a radiance residual field. This models the appearance discrepancies between the renderings of the rigid branch and the ground-truth colors, typically caused by the complex non-rigid motions or deformations.
We additionally condition the radiance residual predictions with the canonical radiance representation of the rigid branch as prior information. This enables more coherent modeling of non-rigid motions, as the non-rigid and rigid radiance fields are highly correlated. 

Specifically, unlike the rigid branch, the queried feature $f_{\mathbf{x}_c}$ is not sliced, enabling the non-rigid branch to learn the radiance residuals with reference to the canonical representation.
Instead, we freeze the first half of $f_{\mathbf{x}_c}$ that represents the canonical radiance field, and only optimize the other half to learn the radiance residuals for non-rigid motion modeling. We denote this partially frozen feature as $f^*_{\mathbf{x}_c}$. We observe that providing frozen $f^{:H}_{\mathbf{x}_c}$ to the non-rigid radiance branch enables more effective and coherent residual learning. 
In addition, we also condition the radiance residual field prediction with our pose-embedded implicit feature $\mathbf{P}_{nr}$, which is achieved by concatenating $\mathbf{P}_{nr}$ to the first intermediate output of the MLP decoder $\mathbf{M}_{nr}$, as illustrated in Fig.~\ref{fig:overview}. The generation of $\mathbf{P}_{nr}$ is explained in the last subsection of Sec.~\ref{sec:method}. Eventually, the color and density residuals are predicted by $\mathbf{M}_{nr}$ for the final compositional radiance computation:
\begin{equation}
    \Delta c,\ \Delta\sigma = \mathbf{M}_{nr}(f^*_{\mathbf{x}_c}; \mathbf{P}_{nr}), 
\end{equation}

\begin{figure*}[t]
    \centering
    \resizebox{1\linewidth}{!}{
        \begin{minipage}{.44\linewidth}
            \centering
            \includegraphics[width=\linewidth]{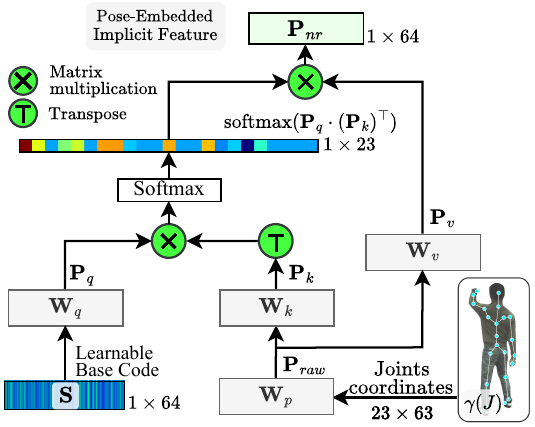}
            \captionof{figure}{Illustration of the proposed pose-embedded implicit feature generation. We employ cross-attention to modulate the single learnable base code to pose-adaptive features.} 
            \label{fig:attention}
        \end{minipage}
        \hspace{.02\linewidth}
        \begin{minipage}{.55\linewidth}
            \centering
            \includegraphics[width=\linewidth]{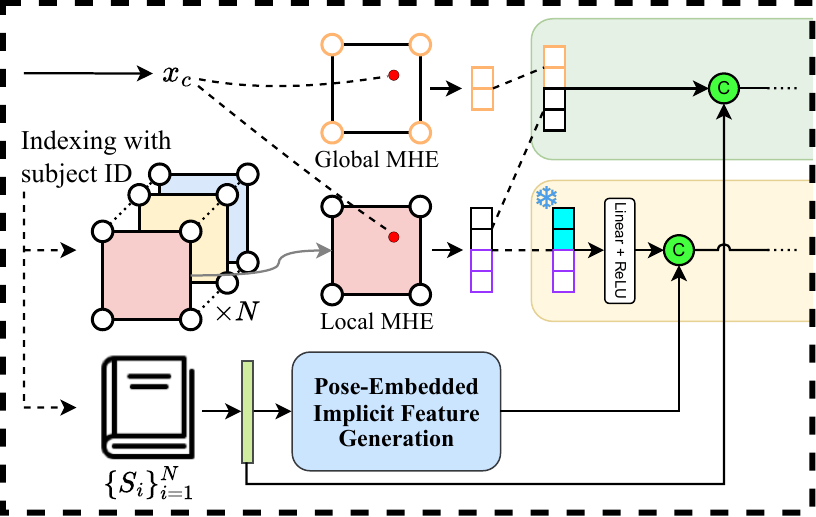}
            \captionof{figure}{Illustration of extended architecture of MoCo-NeRF for the multi-subject unified training (Sec.~\ref{method:multiSubjects}). Major components consist of the global MHE, set of local MHEs, and the dictionary of learnable base codes as ID codes.}
            \label{fig:multiID}
        \end{minipage}
    }
\end{figure*}

\noindent
\textbf{Compositional volume rendering.}
Following \cite{mildenhall2020nerf}, we adopt stratified sampling and volume rendering to predict the RGB color of each ray.
For final rendering, we query $M$ samples per ray $\textbf{r}$ and perform approximated integration to obtain final RGB $C_{final}$ from $(c+\Delta c, \sigma+\Delta\sigma)$ as follows:
\begin{equation}
\begin{split}
    C_{final}(\textbf{r})  = 
    \textstyle\sum_{m=1}^M T_m(1-\text{exp}(-(\sigma_{m}+\Delta\sigma_{m})\delta_{m}))(c_m+\Delta c_m). \label{eq:volumeRender}
\end{split}
\end{equation}
\noindent
where $\delta_{m}$ is the adjacent distance from the $m$-th to the $m+1$-th sample and $T_m=\text{exp}(-\textstyle\sum_{n=1}^{m-1} \sigma_{n} \delta_{n})$ as the transmittance. In addition, we compute $C_{rigid}$ in the same way from $(c, \sigma)$ for an additional loss term that encourages a decomposed learning of the canonical radiance field and the radiance residuals (Sec.~\ref{method:optim}).

\noindent
\textbf{Learnable Pose-Embedded Implicit Feature.}\label{method:poseCodes}
We introduce high-frequency modulated implicit features computed from a learnable base code and a 3D body pose. 
A learnable base code $\mathbf{S}\in\mathbb{R}^{64}$ is randomly initialized and jointly optimized for a single subject with our motion-oriented compositional radiance fields.
To make $\mathbf{S}$ distinguishable for each pose, we propose to employ the 3D body pose as a guidance for computing implicit features that encourage high-fidelity and pose-dependent radiance residual learning.~A cross-attention mechanism is designed to perform interactions between the learnable base code and 23 joint positions (\ie~$J = \{j_i\}^{23}_{i=1}$). A detailed overview of the mechanism is illustrated in Fig.~\ref{fig:attention}. To allow high-frequency implicit feature learning with 3D body joints, we conduct positional encoding~\cite{mildenhall2020nerf} for the input joint coordinates. We first project each joint position into a higher dimension with a sinusoidal positional encoding, \small$\gamma(j_i) = \left(j_i, \sin(2^0 \pi j_i), \cos(2^0 \pi j_i), ..., \sin(2^{L-1} \pi j_i), \cos(2^{L-1} \pi j_i)\right)$\normalsize, where $L=10$ is the number of frequency bands. We then project the encoded points (\ie, $\gamma(J)\in \mathbb{R}^{23\times 63}$) to a raw pose-embedded code $\mathbf{P}_{raw}$ by a projection matrix $\mathbf{W}_p$:
\begin{equation}
    \mathbf{P}_{raw} = \mathbf{W}_p \cdot \gamma(J). \label{eq:3}
\end{equation}
Then, we generate a query signal $\mathbf{P}_q$ from $\mathbf{S}$, and key and value signals from $\mathbf{P}_{raw}$ via linear projections as follows:
\begin{equation}
    \mathbf{P}_q = \mathbf{W}_Q \cdot \mathbf{S},\ \ 
    \mathbf{P}_k = \mathbf{W}_K \cdot \mathbf{P}_{raw},\ \ 
    \mathbf{P}_v = \mathbf{W}_V \cdot \mathbf{P}_{raw}. 
\end{equation}
\noindent
Finally, we generate the learnable pose-embedded implicit feature $\mathbf{P}_{nr}$ for the radiance residual estimation as follows:
\begin{equation}
    \mathbf{P}_{nr} = \text{softmax}(\mathbf{P}_q \cdot (\mathbf{P}_k)^\top) \cdot \mathbf{P}_v. 
\end{equation}
As explained earlier, we employ $\mathbf{P}_{nr}$ to condition $\mathbf{M}_{nr}$, the MLP decoder of the non-rigid radiance branch. $\mathbf{M}_{nr}$ optimizes the implicit feature $\mathbf{P}_{nr}$ and its generation module for estimating pose-coherent radiance residuals instead of solely relying on explicit volumetric features.


\subsection{Multi-subject Unified Training}\label{method:multiSubjects}
Our effective radiance compositional approach for non-rigid motion modeling further enables a flexible extension of our single-subject MoCo-NeRF for unified training with multiple subjects, which is very beneficial for practical applications. 
Each monocular video exhibits distinct non-rigid motions, thereby requiring individualized modeling.
However, it is impossible for the traditional spatial offset learning approach~\cite{weng2022humannerf} that typically requires a heavy module for photo-realistic quality.
In contrast, our radiance compositional approach can achieve it by employing a separate, lightweight MHE for each subject, referred to as local MHE.

Building on this foundation, we propose three architectural modifications as illustrated in Fig.~\ref{fig:multiID}. 
To benefit from unified training of multiple subjects, we introduce a coarse global MHE $\psi^G_{hash}$ shared across all subjects to learn subject-generic representations (\eg, body shapes).
We also ensure learning of subject-specific details with $N_s$ local MHEs $\{\psi^i_{hash}\}^{N_s}_{i=1}$, where $N_s$ is the number of subjects.
For each canonical sample $\mathbf{x}_c$, we query global features from the global MHE and concatenate it to the sliced feature from the corresponding local MHE for radiance predictions. 
We additionally propose to leverage the learnable base codes of all subjects $\{\mathbf{S}_i\}^{N_s}_{i=1}$ as ID codes, for sharing MLP decoders while ensuring subject-distinctive canonical representations.
The modified radiance estimations of the rigid branch for the multi-subject training are as follows:
\begin{equation}
    c,\ \sigma = M_{rigid}(\psi^G_{hash}(\mathbf{x}_c) \oplus (\psi^{i}_{hash}(\mathbf{x}_c))^{:H} \oplus \mathbf{S}_i)
\end{equation}

\subsection{Model Optimization}\label{method:optim}
We optimize MoCo-NeRF with all $N_i$ image frames $\{I_j\}^{N_i}_{j=1}$ from the monocular video via patch-based ray sampling. Given an image frame $I_j$ at each iteration, we randomly sample $N_p$ patches denoted as $\{P_k\}^{N_p}_{k=1}$ to construct the LPIPS~\cite{zhang2018perceptual} loss. 
The LPIPS loss term can measure the perceptual distance between two image patches, and thus we also adopt it for more perceptually coherent renderings.
Our framework first renders $\{\hat{P}^r_k,\hat{P}^f_k\}^{N_p}_{k=1}$ respectively from $C_{rigid}$ and $C_{final}$, and compute two LPIPS loss terms $\mathcal{L}^r_{\text{LPIPS}}$ and $ \mathcal{L}^f_{\text{LPIPS}}$ against the ground truth patch $P^{GT}_k$ by using the pre-trained VGGNet~\cite{simonyan2015vggnet} to  as $\mathcal{L}_\text{LPIPS} = \text{LPIPS}(VGG(\hat{P_k}),\ VGG(P^{GT}_k))$.
We respectively compute the MSE loss terms~$\mathcal{L}^r_{\text{MSE}},\mathcal{L}^f_{\text{MSE}}$ from each of the estimated patches $\{\hat{P}^r_k,\hat{P}^f_k\}^{N_p}_{k=1}$.
Moreover, we additionally compute an appearance matching loss with SSIM $\mathcal{L}^f_{\text{SSIM}}$~\cite{clement2017monodepth} only for the final RGB $\{\hat{P}^f_k\}^{N_p}_{k=1}$, to improve the rendering quality.
In summary, there are 5 loss terms in total: $\mathcal{L}^r_{\text{LPIPS}},\mathcal{L}^r_{\text{MSE}}$ for the canonical radiance field of the rigid branch, and $\mathcal{L}^f_{\text{LPIPS}}, \mathcal{L}^f_{\text{MSE}}, \mathcal{L}^f_{\text{SSIM}}$ for the final motion-oriented compositional renderings.
To enforce the rigid branch to learn the main canonical radiance fields while the non-rigid branch to learn radiance residuals, we compute the final overall loss function~$\mathcal{L}^o$ with a hyperparameter $\lambda=0.2$ as:
\begin{gather}
        \mathcal{L}^o = \lambda\mathcal{L}^r + (1-\lambda)\mathcal{L}^f \\
    \mathcal{L}^r = \mathcal{L}^r_{\text{LPIPS}} + \mathcal{L}^r_{\text{MSE}}, \hspace{0.5em}
    \mathcal{L}^f = \mathcal{L}^f_{\text{LPIPS}} + \mathcal{L}^f_{\text{MSE}} + \mathcal{L}^f_{\text{SSIM}}
\end{gather}

\section{Experiments}
We conduct extensive experiments on ZJU-MoCap~\cite{peng2021neuralbody} and MonoCap~\cite{peng2021animatable, habermann2021, habermann2020deepcap} to verify the effectiveness of our proposed approach on both single- and multi-subject settings. The multi-subject setting indicates multiple subjects are jointly optimized. Our findings demonstrate that MoCo-NeRF can learn photo-realistic dynamic human representations by effectively handling non-rigid motions. Moreover, it learns discriminative representations of each performer in the multi-subject setting, while prior works struggle.

\subsection{Datasets}
We leverage ZJU-MoCap and MonoCap datasets for both training and evaluation, following~\cite{geng2023instantnvr, hu2024gauhuman}, and provide the details in this section. For ZJU-MoCap, we evaluate our MoCo-NeRF and other baselines on all 6 subjects~(\ie, 377, 386, 387, 392, 393, 394). Out of 23 camera views, we use one of them for training and the other 22 views for evaluation. For MonoCap that comprises DynaCap~\cite{habermann2021} and DeepCap~\cite{habermann2020deepcap}, we utilize all 4 subjects~(\ie, lan, marc, olek, vlad) and use the same set of cameras as~\cite{geng2023instantnvr}. For multi-subject evaluations presented in Tab.~\ref{tab:quant_multi}, we evaluate on subjects 377, 386, and 392 of ZJU-MoCap.

\begin{table}[!t]
    \centering
    \caption{Quantitative results of single-subject methods on ZJU-MoCap and MonoCap datasets where LPIPS$^*$ $=$ LPIPS$\times 10^3$ with the \textbf{best} and \underline{second-best} performance notations. 
    MoCo-NeRF outperforms the baselines on most of the evaluation metrics in both settings. Despite pixel-wise metrics (PSNR, SSIM) tend to favor smooth renderings generated from \cite{geng2023instantnvr, hu2024gauhuman}, we achieve state-of-the-art performance while effectively modeling the photo-realistic non-rigid motions as illustrated in Fig.~\ref{fig:qualitative}.}
    \footnotesize
    \setlength{\tabcolsep}{5pt}
    \begin{tabular}{ l | c c c | c c c}
        \toprule
        \multicolumn{1}{c|}{Single-subject} & \multicolumn{3}{c|}{ZJU-MoCap~\cite{peng2021neuralbody}} & \multicolumn{3}{c}{MonoCap~\cite{peng2021animatable}} \\
        
        \multicolumn{1}{c|}{methods} & PSNR\hspace{1pt}$\uparrow$ & SSIM\hspace{1pt}$\uparrow$ & LPIPS$^*$$\downarrow$ & PSNR\hspace{1pt}$\uparrow$ & SSIM\hspace{1pt}$\uparrow$ & LPIPS$^*$$\downarrow$ \\ \midrule 

        HumanNeRF~\cite{weng2022humannerf} 
            & 30.18 & \underline{0.9709} & \underline{30.13} & \underline{32.92} & 0.9864 & \textbf{12.93} \\

        Instant-NVR~\cite{geng2023instantnvr}
            & 30.73 & 0.9697 & 38.87 & 31.89 & \underline{0.9870} & 16.07 \\

        GauHuman~\cite{hu2024gauhuman} 
            & \underline{30.98} & 0.9620 & 30.16& 32.87 & 0.9849 & 13.54 \\

        MoCo-NeRF (Ours)
            & \textbf{31.06} & \textbf{0.9734} & \textbf{28.83} & \textbf{33.01} & \textbf{0.9874} & \underline{13.02} \\
        \bottomrule
    \end{tabular}
    
    \label{tab:quant_single}
\end{table}

\begin{figure*}[!t]
\centering
  \includegraphics[width=1\linewidth]{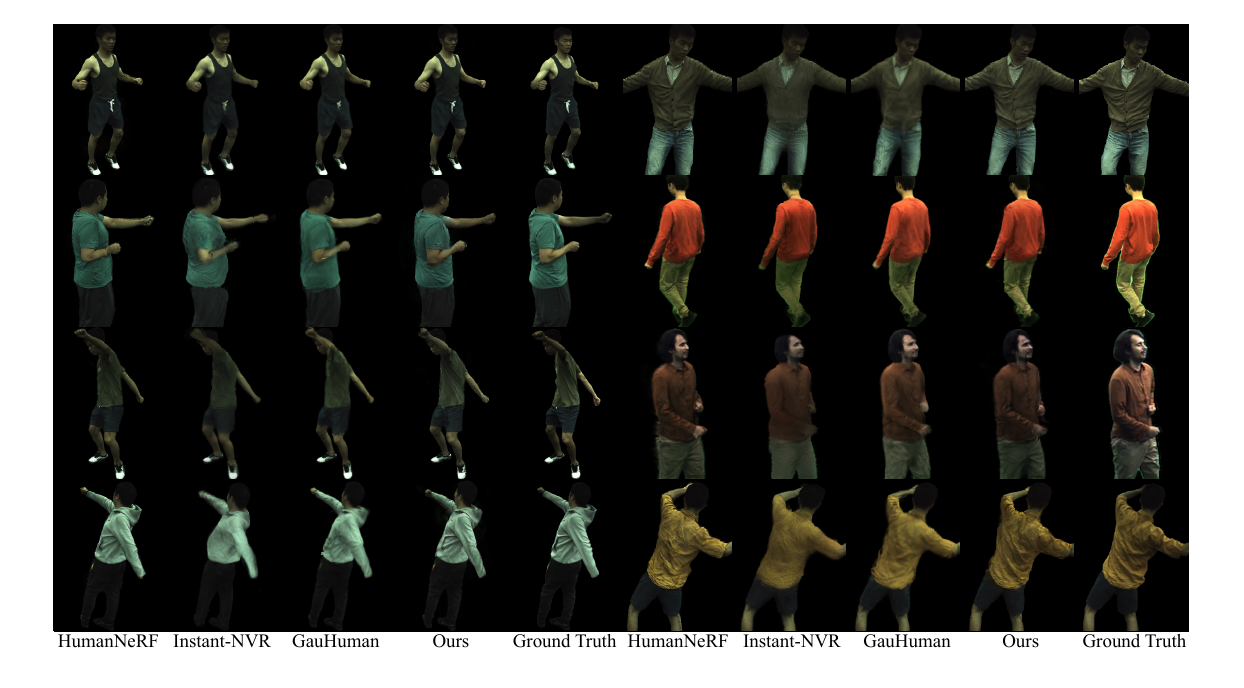}%
  \caption{Qualitative comparison of novel view synthesis from single-subject training. MoCo-NeRF achieves photo-realistic rendering qualities with high-fidelity non-rigid motion modeling. Although HumanNeRF~\cite{weng2022humannerf} presents comparable quality, however, MoCo-NeRF achieves much less training time as shown in Tab.~\ref{tab:efficiency}.}
  \label{fig:qualitative}
\end{figure*}

\subsection{Implementation Details}
We optimize our model with Adam optimizer~\cite{kingma2015adam} with betas $(0.9, 0.99)$, and learning rates of $5e^{-5}$ for the skeletal weights volume decoder and $5e^{-3}$ for the others. Each MHE has 16 levels with feature dimensions $H=2$, and 4 levels with $H=2$ for the global MHE. For each iteration, six $32\times32$ patches are used with 256 samples per ray. We train all models including the baselines on 1 Nvidia RTX 3090, where both baselines are trained for their suggested duration. Our model is trained for $1\sim2$ hours depending on each subject's complexity for the single-subject training, and $2$ hours for the multi-subject training.

\begin{table}[t]
    \centering
    \caption{Quantitative results of multi-subject methods on subjects \#377, \#386, \#392 of ZJU-MoCap~\cite{peng2021neuralbody}. 
    Despite the same evaluation settings, MoCo-NeRF significantly outperforms baseline methods under monocular settings with a shorter training time.}
    \footnotesize
    \setlength{\tabcolsep}{4pt}
    \begin{tabular}{ l|ccccc }
        \toprule
        \multicolumn{1}{c|}{Multi-subject methods} & PSNR\hspace{1pt}$\uparrow$ & SSIM\hspace{1pt}$\uparrow$ & LPIPS$^*$$\downarrow$ & Training videos & Train time \\ \midrule 
        NHP~\cite{kwon2021neural} & 26.64 & 0.9185 & 40.47 & multi-view & {\scriptsize$\sim$}86 hours \\ 
        GHuNeRF~\cite{li2023ghunerf} & 27.28 & 0.9291 & 40.82 & \textbf{monocular} & {\scriptsize$\sim$}123 hours \\ 
        MoCo-NeRF (Ours) & \textbf{31.97} & \textbf{0.9781} & \textbf{24.54} & \textbf{monocular} & \textbf{{\scriptsize$\sim$}2 hours} \\ 

        \bottomrule
    \end{tabular}
    \normalsize
    \label{tab:quant_multi}
\end{table}

\begin{figure*}[!t]
\centering
  \includegraphics[width=1\linewidth]{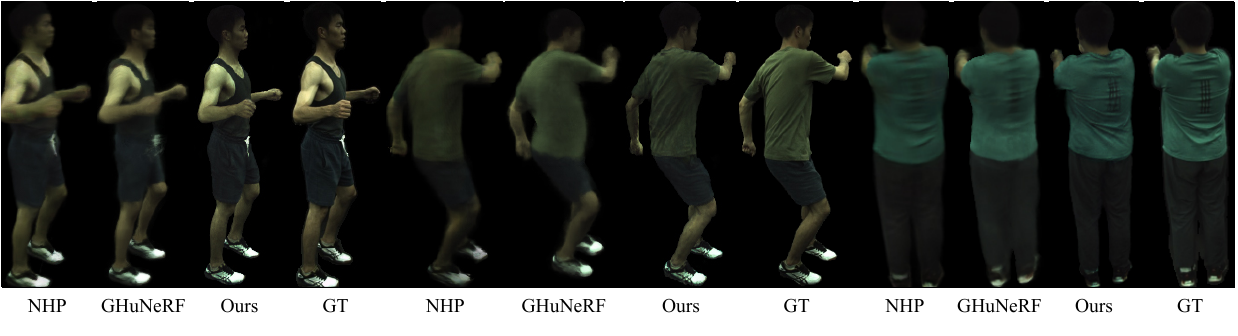}%
  \caption{Qualitative comparison of novel view synthesis from multi-subject training. Although all subjects are seen during training, only our multi-subject MoCo-NeRF achieves photo-realistic quality comparable to the ones of single-subject models.}
  \label{fig:qualitative_multi}
\end{figure*}

\subsection{State-of-the-art Comparison}
\textbf{Single-subject methods.}
Tab.~\ref{tab:quant_single} presents PSNR, SSIM~\cite{wang2004ssim}, and LPIPS$^*$ of MoCo-NeRF and our baselines~\cite{weng2022humannerf, geng2023instantnvr, hu2024gauhuman}, for novel-view synthesis under the single-subject setting. MoCo-NeRF achieves state-of-the-art performance on all metrics except one comparable LPIPS$^*$ for the MonoCap dataset. As shown in Fig.~\ref{fig:qualitative}, Instant-NVR~\cite{geng2023instantnvr} shows moderate performance, however, it often generates smoothed-out results without proper non-rigid motions due to high dependency on accurate SMPL parameters. GauHuman~\cite{hu2024gauhuman} achieves better qualitative performance than~\cite{geng2023instantnvr} but still struggles to model sophisticated non-rigid motions. Fig.~\ref{fig:teaser} presents longer optimizations are surprisingly ineffective and harmful respectively for Instant-NVR~\cite{geng2023instantnvr} and GauHuman~\cite{hu2024gauhuman} in terms of non-rigid motion modeling.
This implies that their approaches cannot achieve optimal non-rigid motion modeling.
Moreover, we observe that PSNR and SSIM often favor those results without fine-grained non-rigid motion modeling since predicting pose-dependent deformations from novel viewpoints may not be pixel-wise accurate. Instead, the perception metric (LPIPS$^*$) evaluates better in this case, as HumanNeRF~\cite{weng2022humannerf} has a better LPIPS$^*$ score and also shows qualitatively better results. Thus, we consider GauHuman's comparable PSNR might be favored by a superior representation (3D Gaussians~\cite{kerbl3Dgaussians}) instead of better non-rigid motion modeling. A more detailed analysis is in supplementary materials. Despite this, we still achieve state-of-the-art performance with significantly improved training efficiency compared to another photo-realistic model~\cite{weng2022humannerf} (Tab.~\ref{tab:efficiency}). 

\begin{table}[t]
    \centering
    \caption{Training efficiency comparison for methods with photo-realistic non-rigid motion modeling on a single RTX3090. In the table, \textit{single-sub.} and \textit{multi-sub.} respectively represent training for 1 and 6 subjects. MoCo-NeRF significantly outperforms HumanNeRF in terms of both GPU memory usage and training time. HumanNeRF requires sequential training to achieve comparable non-rigid motion modeling for all subjects, thus resulting in linearly-increased training time.}
    \footnotesize
    \setlength{\tabcolsep}{2.5pt}
    \begin{tabular}{ l|ccc|ccc }
        \toprule
        & \multicolumn{3}{c|}{GPU Mem. (GB)} & \multicolumn{3}{c}{Train time (hours)} \\ 
        & \textit{single-sub.} & \textit{multi-sub.} & \multicolumn{1}{c|}{$\Delta$} & \textit{single-sub.} & \textit{multi-sub.} & $\Delta$ \\ \midrule 
        HumanNeRF~\cite{weng2022humannerf} & 23.7 & 23.7 & (+0) & 38.9 & 233.3 &\textcolor{red}{(+194.5)} \\ 
        MoCo-NeRF (Ours) & \textbf{16.2} & \textbf{20.3} & (+4.1) & \textbf{1.9} & \textbf{2.1} &\textbf{(+0.2)} \\ 

        \bottomrule
    \end{tabular}
    \normalsize
    
    \label{tab:efficiency}
\end{table}

\noindent
\textbf{Multi-subject methods.}
We evaluate our MoCo-NeRF against NHP~\cite{kwon2021neural} and GHuNeRF~\cite{li2023ghunerf}, both of which are designed to achieve generalization capabilities. This is necessitated by the absence of prior works that directly target the joint learning task for multiple subjects from \textit{monocular videos}. Additionally, we observe that the single-subject baseline models fail to learn distinct representations when trained in a unified manner. This observation underscores our decision to compare against generalization models that are specifically tailored to leverage multi-subject data for training. For fairness, we use their official checkpoints and ensure all evaluated subjects were included for training.

As demonstrated in Tab.~\ref{tab:quant_multi}, MoCo-NeRF remarkably outperforms all the baseline methods on all metrics while also excelling in training efficiency up to 61 times faster. Even though MoCo-NeRF only utilizes a monocular video of each subject, Fig.~\ref{fig:qualitative_multi} illustrates our method succeeds in learning both distinct representations and pose-dependent radiance residuals of each subject, enabling photo-realistic free-viewpoint renderings.

\subsection{Model Analysis}\label{sec:ablation}
In this section, we present analyses of our contributions respectively for two major achievements: novel non-rigid motion modeling (\textbf{NR modeling}) and multi-subject unified training (\textbf{multi-subject modeling}). For both analyses, we progressively remove each component from a full model to a vanilla approach.

\begin{figure*}[!t]
\centering
  \includegraphics[width=1\linewidth]{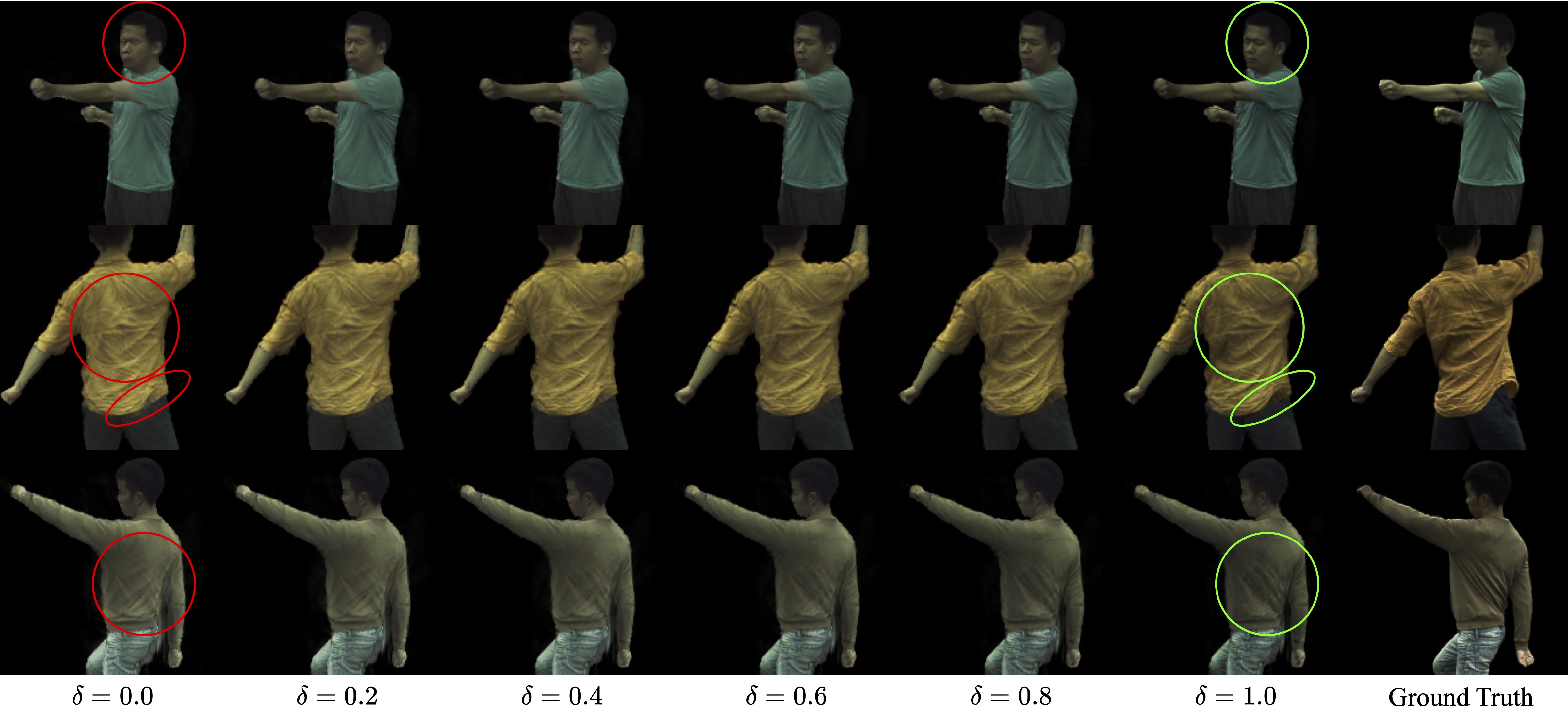}%
  \caption{Qualitative illustrations of non-rigid radiance residuals on novel-view synthesis where $\delta \rightarrow c' = c +\delta\Delta c, \sigma'=\sigma + \delta\Delta\sigma$. The first row progressively demonstrates the capability of the non-rigid radiance branch to learn the non-rigid motions on the subject's face in addition to the cloth folds. The last two rows clearly illustrate the effects of radiance residuals on cloth folds and hems corresponding to the body pose.}
  \label{fig:progressive_residual}
\end{figure*}

\noindent
\textbf{NR Modeling: Analysis of radiance compositional approach.}~Tab.~\ref{tab:ablation_single} demonstrates our radiance compositional design perceptually improves the rendering results by closing the radiance discrepancies with predicted residuals. 
Fig.~\ref{fig:ablation_single} shows that the representation learned solely by the rigid radiance branch (c) suffers from noisy renderings caused by incorrect correspondences due to non-rigid motions.
With our radiance residual modeling (b), those issues are mitigated by jointly learning per-sample radiance residuals, resulting in more complete rendering and better non-rigid motion modeling as clearly shown in Fig.~\ref{fig:ablation_single}.
In addition, Fig.~\ref{fig:progressive_residual} illustrates successful decomposed learning of two types of representations and validates the effectiveness of our radiance compositional approach in non-rigid motion modeling.

\noindent
\textbf{NR Modeling: Analysis of pose-embedded implicit feature.}
Tab.~\ref{tab:ablation_single} further illustrates that the final version with the proposed pose-embedded implicit features (Tab.~\ref{tab:ablation_single} (a)) significantly improves the quantitative performance on all metrics.
It enables the non-rigid branch to implicitly utilize body pose information in addition to the explicit features, yielding smoother and more coherent non-rigid motion modeling as shown in Fig.~\ref{fig:ablation_single}. We can observe remarkable quality enhancements that demonstrate the effectiveness of our attention-based code generation module. We also present an ablation study on the choice of cross-attention module for the feature generation in the supplementary material.

\begin{table*}[t]
    \centering
    \begin{minipage}[t]{\textwidth}
    \begin{minipage}{.49\linewidth}
        \centering
        \captionof{table}{Ablation study in the single-subject task on ZJU-MoCap for b) pose-embedded implicit features and c) motion-oriented compositional NeRF approach.}
        \scriptsize
        \begin{tabular}{ l l | c c c }
            \toprule
            \multicolumn{2}{c|}{Single-subject} & PSNR & SSIM & LPIPS$^*$ \\ \midrule 
            \multicolumn{1}{c}{(a)} & MoCo-NeRF (Ours) & \textbf{31.06} & \textbf{0.9734} & \textbf{28.83} \\
            \multicolumn{1}{c}{(b)} & (--) Pose features & 30.64 & 0.9702 & 29.65 \\
            \multicolumn{1}{c}{(c)} & (--) Residual NeRF & 30.62 & 0.9702 & 29.93 \\
    
            \bottomrule
        \end{tabular}
        \normalsize
        \label{tab:ablation_single}
    
    \end{minipage}
    \hfill
    \begin{minipage}{.49\linewidth}    
    \centering
    \captionof{table}{Ablation study of multi-subject MoCo-NeRF on ZJU-MoCap for our proposed architectural modifications.}
    \scriptsize
    \setlength{\tabcolsep}{1.3pt}
    \begin{tabular}{ l l | c c c }
        \toprule
        \multicolumn{2}{c|}{Multi-subject} & PSNR & SSIM & LPIPS$^*$ \\ \midrule
        
        \multicolumn{1}{c}{(i)} & MoCo-NeRF (Ours) & \textbf{30.91} & \textbf{0.9746} & \textbf{27.69} \\
        \multicolumn{1}{c}{(ii)} & (--) Global MHE & 30.83 & \textbf{0.9746} & 28.49 \\
        \multicolumn{1}{c}{(iii)} & (--) $N_s$ local MHEs & 30.70 & 0.9742 & 29.03 \\
        \multicolumn{1}{c}{(iv)} & (--) Identity codes & 28.63 & 0.9666 & 38.29 \\

        \bottomrule
    \end{tabular}
    \normalsize
    \label{tab:ablation_multi}
    \end{minipage}
    \end{minipage}
\end{table*}
\begin{table*}[t]
    \centering
    \small
    \begin{minipage}{\textwidth}
    \begin{minipage}[]{.49\linewidth}

        \centering
        \includegraphics[width=\linewidth]{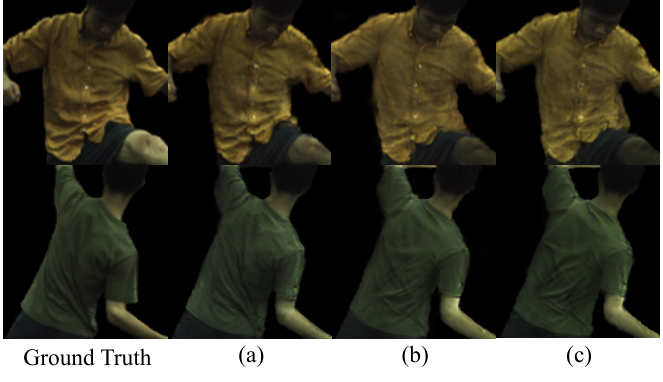}

    \end{minipage}
    \hfill
    \begin{minipage}[c]{.49\linewidth}    
    \centering
    \includegraphics[width=\linewidth]{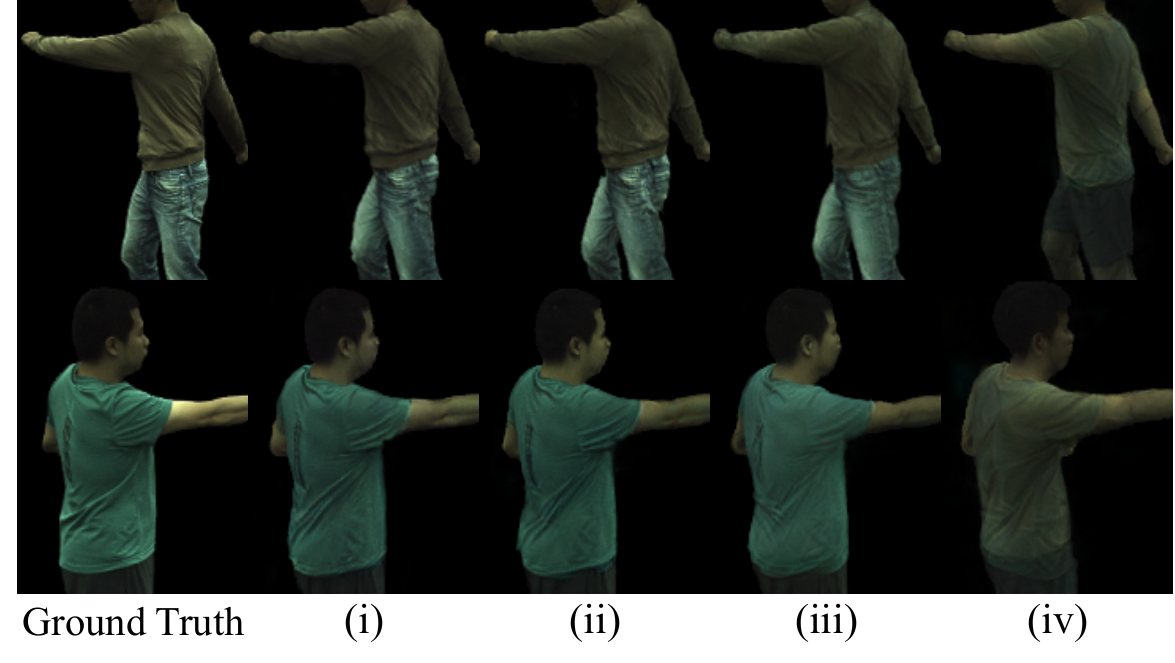}
    
    \end{minipage}
    
    \begin{minipage}[t]{.49\linewidth}
        \captionof{figure}{Qualitative results of the ablation study on single-subject MoCo-NeRF presented in Tab.~\ref{tab:ablation_single}.}
        \label{fig:ablation_single}
    \end{minipage}
    \hfill
    \begin{minipage}[t]{.49\linewidth}
        \captionof{figure}{Qualitative results of the ablation study in Tab.~\ref{tab:ablation_multi} where we can observe gradual improvements from (iv) to (i).}
        \label{fig:ablation_multi}
    \end{minipage}
    \end{minipage}
\end{table*}

\noindent
\textbf{Multi-subject Modeling: Analysis of using learnable base codes as ID codes.}
In Fig.~\ref{fig:ablation_multi}, the results of MoCo-NeRF without any modifications (iv) exhibit mixed-up appearances on all target subjects, similar to the ones of~\cite{weng2022humannerf} that are presented in the supplementary material. Utilizing a set of learnable base codes as ID codes (Tab.~\ref{tab:ablation_multi} (iii)) enables a single MLP decoder $M_{rigid}$ to differentiate each subject by optimizing the ID codes as plug-in parameters. Thus, distinct representations of all subjects are successfully learned as illustrated in (iii) of Fig.~\ref{fig:ablation_multi}, yielding significant quantitative improvements as well (Tab.~\ref{tab:ablation_multi}).  

\noindent
\textbf{Multi-subject Modeling: Analysis of $N_s$ local MHEs.}
Fig.~\ref{fig:ablation_multi} shows that the model (iii) lacks fine-grained details due to the sharing of one MHE. By utilizing separate $N_s$ local MHEs (Tab.~\ref{tab:ablation_multi} (ii)), the model learns subject-discriminative non-rigid motions as they now optimize separate explicit features for each subject's radiance residuals. Fig.~\ref{fig:ablation_multi} shows clear improvement of the model (ii) in non-rigid motion modeling where its quantitative gains are reflected in Tab.~\ref{tab:ablation_multi}.

\noindent
\textbf{Multi-subject Modeling: Analysis of global MHE.}
The addition of global MHE (i) further empowers MoCo-NeRF to learn better representations with more coherent non-rigid motions as illustrated in Fig.~\ref{fig:ablation_multi}. 
Although the global MHE is not used for learning non-rigid radiance residuals, it encourages the rigid part of features from local MHE to concentrate on subject-specific representations.
As its frozen copy is used for the non-rigid radiance branch, it facilitates better learning of radiance residuals for modeling non-rigid motions.~Tab.~\ref{tab:ablation_multi} clearly shows that it can help to learn perceptually better representations.

\section{Conclusion}
In this paper, we presented a novel framework MoCo-NeRF for free-viewpoint dynamic human rendering that is capable of both single- and multi-subject training from monocular videos. By decomposing the neural representation of a dynamic human by motions and utilizing proposed implicit pose codes, MoCo-NeRF effectively learns radiance discrepancies induced by non-rigid motions. The proposed multi-subject MoCo-NeRF efficiently achieves comparable performance compared to the single-subject model. It has fully demonstrated its effectiveness and established new state-of-the-art performance on both problems. \\
\textbf{Acknowledgments.} This research is supported in part by the Hong Kong PhD Fellowship Scheme (HKPFS), Early Career Scheme of the Research Grants Council (RGC) of the Hong Kong SAR under grant No. 26202321, SAIL Research Project, HKUST-Zeekr Collaborative Research Fund, HKUST-WeBank Joint Lab Project, Tencent Rhino-Bird Focused Research Program, and NAVER Cloud Corporation.

%
%
\bibliographystyle{splncs04}
\bibliography{main}


\renewcommand{\thesection}{\Alph{section}.}
\renewcommand{\thesubsection}{\thesection\arabic{subsection}}
\renewcommand\thefigure{\Alph{figure}}
\renewcommand\thetable{\Alph{table}}
\setcounter{section}{0}
\setcounter{figure}{0}
\setcounter{table}{0}

\section{Comparisons with GauHuman (Effectiveness of Pixel-Wise Metrics)}
\begin{figure}
    \centering
    \includegraphics[width=.99\textwidth]{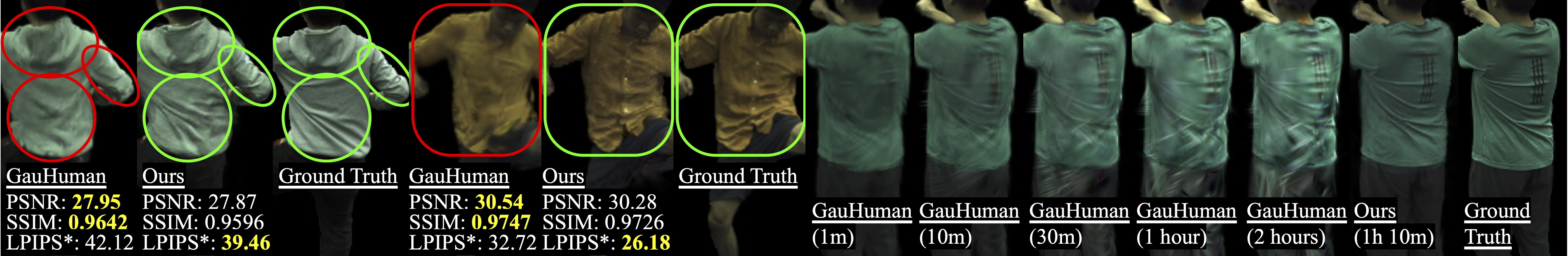} 
    \caption{Failures of pixel-wise metrics and longer training of GauHuman~\cite{hu2024gauhuman}.}
    \label{fig:lpips_gauhuman}
\end{figure}

In the main paper, we mention that the pixel-wise metrics (PSNR, SSIM) favor smoother renderings generated by GauHuman~\cite{hu2024gauhuman}, which is a concurrent work published one week prior to our submission.
As results, our proposed MoCo-NeRF and GauHuman show comparable PSNR on both datasets and we would like to provide an analysis on this phenomenon.
As mentioned in the main paper, GauHuman cannot capture \textit{pose-dependent cloth wrinkles} as effectively as our model. Pixel-wise metrics (PSNR) are less sensitive to these details, but better reflected by perceptual metric LPIPS$^*$ as shown in Fig.~\ref{fig:lpips_gauhuman}. Nevertheless, Tab.~1 in the paper shows our significant improvements in \textbf{LPIPS$^*$}and SSIM. Fig.~5 in the paper and Fig.~\ref{fig:lpips_gauhuman} show GauHuman \textit{\textbf{lacks detailed cloth wrinkles}} for \underline{most of subjects}, even with longer training. This limitation may be due to its reliance on accurate parametric model (SMPL) like Instant-NVR, while ours is free from the dependency. Overall, our work achieves clear state-of-the-art in single-subject task as well.

\section{Integration of MHE into HumanNeRF}
\begin{figure}[!t]
    \centering
    \includegraphics[width=.99\textwidth]{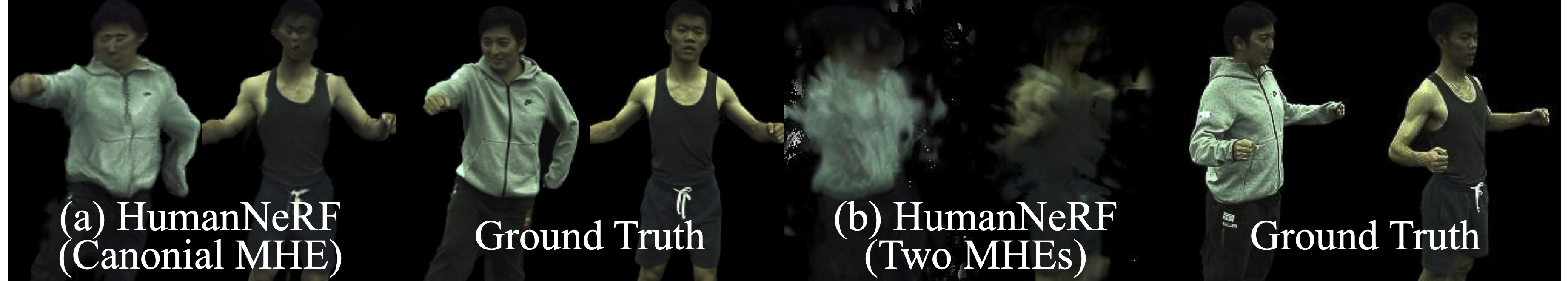} 
    \caption{Illustrations of HumanNeRF with MHEs.}
    \label{fig:humannerf_mhe}
\end{figure}

In this section, we show that replacing MLPs of HumanNeRF~\cite{weng2022humannerf} with MHEs is non-trivial. 
First, if we replace canonical MLP of HumanNeRF with MHE and train for an hour, it results in wiggly novel-view appearances as shown in Fig.~\ref{fig:humannerf_mhe}~(a). It is due to the non-rigid MLP, which converges in 39 hours, predicting inconsistent spatial offsets $\Delta \mathbf{x}$ for each canonical point $\mathbf{x}_c$. As a result, multiple hash entries are created for the same $\mathbf{x}_c$. 

Fig.~\ref{fig:humannerf_mhe}~(b) shows using both non-rigid and canonical MHEs worsens the results as explicit MHE fails to replicate the implicit MLP's non-rigid motion modeling despite the same training configurations. In contrast, our novel motion modeling enables not only successful modeling of non-rigid motions but also optimal utilization of MHE, yielding faster and better performance.

\section{Additional Ablation Study on Multi-Subject MoCo-NeRF}
Our ablation studies cumulatively removing each component might mislead that the introduction of ID codes contribute the most to the performance, but this is not true.
As shown in Tab.~\ref{tab:ablation_nmhe}, adding $N_s$ local MHEs to the vanilla single-subject MoCo-NeRF (Tab.~5 (iv) in main paper) also significantly boosts the performance, paralleling the impact of ID codes.
Therefore, other components beyond ID codes also greatly contribute to state-of-the-art performance.

\setlength{\tabcolsep}{7pt}
\begin{table}
    \centering
    \caption{Ablation studies on $N$ local MHEs for multi-subject training.}
    \begin{tabular}{ c | c c c }
        \toprule
        & PSNR $\uparrow$ & SSIM $\uparrow$ & LPIPS$^*$ $\downarrow$ \\ \midrule 
        Vanilla Single-subject MoCo-NeRF & 28.63 & 0.9666 & 38.29 \\
        (+) $N$ local MHEs & \textbf{30.86} & \textbf{0.9741} & \textbf{28.80} \\
        \bottomrule
    \end{tabular}
    \normalsize
    \label{tab:ablation_nmhe}
\end{table}
\setlength{\tabcolsep}{6pt} 

\section{Reason for Splitting MHE Features}
Leveraging two separate MHEs for rigid and non-rigid branches of MoCo-NeRF slows down its training and rendering by $\times$1.14 due to an overhead of querying hash tables twice. Our proposed novel feature-splitting approach successfully avoids the overhead by querying the hash tables only once while optimizing independent hash features for two different purposes.

\section{Novel Pose Synthesis}
\begin{figure}
    \centering
    \includegraphics[width=.99\textwidth]{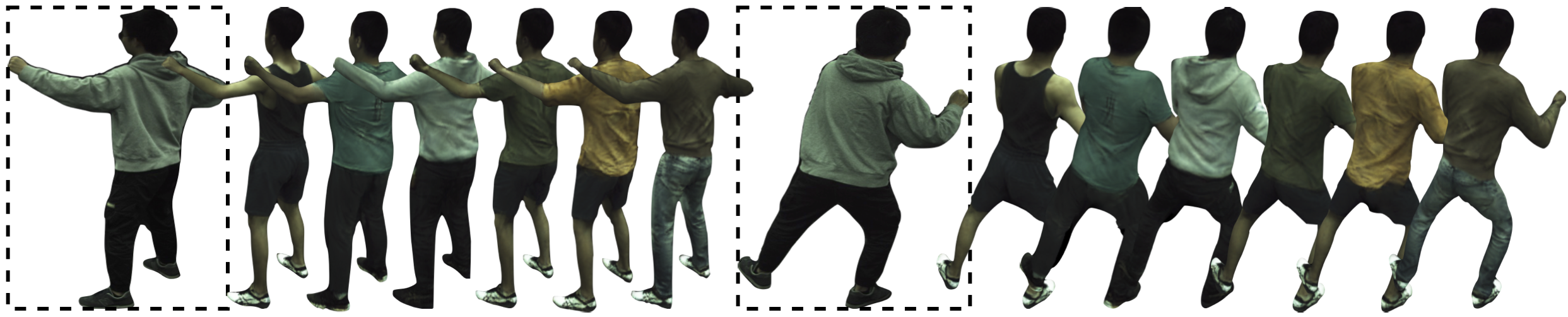} 
    \caption{Illustrations of novel-pose synthesis of MoCo-NeRF.}
    \label{fig:novel_pose}
\end{figure}

Although MoCo-NeRF does not specifically target novel-pose animations, we still can transfer unseen poses to the learned representations as shown in Fig.~\ref{fig:novel_pose}.

\section{Additional Qualitative Comparisons}
In the demo video \emph{\textcolor{red}{demo.mp4}}, we additionally illustrate the effectiveness of MoCo-NeRF in handling non-rigid motions and modeling multiple subjects at the same time. In two sections of the video, we compare against single-subject methods~\cite{weng2022humannerf, geng2023instantnvr, hu2024gauhuman} on both single- and multi-subject training settings, respectively.

In the single-subject setting, MoCo-NeRF achieves photo-realistic novel-view synthesis via motion-based compositional NeRF while GauHuman~\cite{hu2024gauhuman} and Instant-NVR~\cite{geng2023instantnvr} fail to model coherent pose-dependent non-rigid motions. Despite HumanNeRF's comparable quality, its training efficiency is much worse than our proposed MoCo-NeRF as discussed in the main paper. For the multi-subject setting, all single-subject methods~\cite{weng2022humannerf, geng2023instantnvr, hu2024gauhuman} cannot model distinct representations for each subject. On the other hand, MoCo-NeRF readily extends to handle multiple distinct representations benefiting from the radiance compositional approach as also discussed in the main paper.

In the last section of the demo video, we compare multi-subject MoCo-NeRF against GHuNeRF~\cite{li2023ghunerf} and NHP~\cite{kwon2021neural} that are trained with videos of multiple subjects for generalization capability. As mentioned in the main paper, we compare against these models targeting different task since there is no prior work learning distinct representations for multiple subjects from monocular videos. The demo video clearly illustrates that only MoCo-NeRF achieves photo-realistic quality of renderings with pose-dependent non-rigid motions modeling as well. NHP~\cite{kwon2021neural} also presents a slight modeling of non-rigid motions, however, it requires multi-view videos for its training and rendering.

\section{Additional Model Analysis}
\noindent
\textbf{Ablation study of cross-attention for implicit features:} 
In the main paper, we introduce to adopt cross-attention mechanism on a learnable base code $\mathbf{S}$ and input body poses $J$ to memory-efficiently generate pose-adaptive implicit features for the radiance residual field. To analyze its effectiveness, we compare the performance when directly using $\mathbf{S}$ and $J$ as the conditions instead of $\mathbf{P_{nr}}$ in Tab.~\ref{tab:abl_cross}. It demonstrates that using cross-attention to generate pose-coherent conditions can further improves the performance.

\noindent
\textbf{Ablation study of feature sharing methods for our compositional radiance fields:} For our radiance residual predictions, we propose to utilize \emph{frozen features} of the canonical radiance field as representational priors to encourage less complex learning and more coherent non-rigid motion modeling. To analyze the effectiveness of this specific design, we present results when features are not shared (i) and features are shared without freezing (ii) in Tab.~\ref{tab:abl_frozen}. 
Interestingly, allowing shared features to remain unfrozen results in performance degradation, as the learning process for radiance residual fields interferes with that of the canonical radiance field.
Overall, it shows that frozen feature yields the best performance, implying that our motivation to provide representational priors for learning radiance residuals is satisfied.

\begin{table}
    \centering
    \caption{Analysis of cross-attention for pose-embedded implicit feature generation.}
    \begin{tabular}{ c l | c c c }
        \toprule
        \multicolumn{2}{c|}{Single-subject} & PSNR\hspace{1pt}$\uparrow$ & SSIM\hspace{1pt}$\uparrow$ & LPIPS$^*$$\downarrow$ \\ 
        \midrule 
        (a) & Ours (w/o cross-attention) & 30.88 & 0.9706 & 30.66 \\
        (b) & Ours (full model) & \textbf{31.06} & \textbf{0.9734} & \textbf{28.83} \\

        \bottomrule
    \end{tabular}
    \label{tab:abl_cross}
\end{table}

\begin{table}
    \centering
    \caption{Analysis of feature sharing methods for our proposed compositional NeRF architecture.}
    \begin{tabular}{ c l | c c c }
        \toprule
        \multicolumn{2}{c|}{Single-subject} & PSNR\hspace{1pt}$\uparrow$ & SSIM\hspace{1pt}$\uparrow$ & LPIPS$^*$$\downarrow$ \\ 
        \midrule 
        (i) & Ours (w/ no feature sharing) & \underline{30.88} & \underline{0.9717} & \underline{30.17} \\
        (ii) & Ours (w/ unfrozen feature sharing) & 30.76 & 0.9710 & 30.71 \\
        (iii) & Ours (full model) & \textbf{31.06} & \textbf{0.9734} & \textbf{28.83} \\

        \bottomrule
    \end{tabular}
    \label{tab:abl_frozen}
\end{table}

\end{document}